\documentclass[conference]{IEEEtran}
\IEEEoverridecommandlockouts

\usepackage{cite}
\usepackage{amsmath,amssymb,amsfonts}
\usepackage{algorithmic}
\usepackage{graphicx}
\usepackage{textcomp}
\usepackage{xcolor}

\usepackage{csquotes} %
\usepackage{colortbl} %
\usepackage{url}
\usepackage{booktabs}

\usepackage{tikz, hyperref}
\usepackage{eso-pic}
\newcommand\copyrighttext{%
\footnotesize \textcopyright 2026 IEEE. Personal use of this material is permitted. Permission from IEEE must be obtained for all other uses, in any current or future media, including reprinting/republishing this material for advertising or promotional purposes, creating new collective works, for resale or redistribution to servers or lists, or reuse of any copyrighted component of this work in other works.
}
\newcommand\copyrightnotice{%
\begin{tikzpicture}[remember picture,overlay]
\node[anchor=south,yshift=10pt] at (current page.south) {\fbox{\parbox{\dimexpr\textwidth-\fboxsep-\fboxrule\relax}{\copyrighttext}}};
\end{tikzpicture}%
}

\def\BibTeX{{\rm B\kern-.05em{\sc i\kern-.025em b}\kern-.08em
    T\kern-.1667em\lower.7ex\hbox{E}\kern-.125emX}}
\begin{document}

\title{PlayerEmbed: Learning Behavioral Fingerprint Representations from Raw Input Sequences

\thanks{
This work was funded by the Deutsche Forschungsgemeinschaft (DFG, German Research Foundation) under Germany's Excellence Strategy---EXC 2070 Grant No. 390732324 (Cluster of Excellence PhenoRob) and by the State of North Rhine-Westphalia through the Lamarr Institute for Machine Learning and Artificial Intelligence.
}
}

\author{\IEEEauthorblockN{Tobias Schneider\textsuperscript{1,4},
Lorenz Sparrenberg\textsuperscript{3,4},
Priya Tomar\textsuperscript{1,2,3,4},\\
Shahzeb Qamar\textsuperscript{1,2,3,4},
Tobias Deußer\textsuperscript{3,4},
Rafet Sifa\textsuperscript{1,3,4}
\IEEEauthorblockA{\textsuperscript{1}Fraunhofer IAIS, Sankt Augustin, Germany \\
\textsuperscript{2}PhenoRob Cluster of Excellence, University of Bonn, Germany\\
\textsuperscript{3}Lamarr Institute for Machine Learning and Artificial Intelligence, Germany\\
\textsuperscript{4}University of Bonn, Germany \\
}}}

\AddToShipoutPictureFG*{\copyrightnotice}
\maketitle

\begin{abstract}
Raw player input sequences in video games contain fine-grained behavior, but are high-dimensional and noisy, which makes large-scale analytics challenging.
To address this, we introduce PlayerEmbed, a framework that learns compact and discriminative player embeddings using deep metric learning.
Our approach uses a hybrid CNN-LSTM encoder trained on raw input sequences from 1,000 players.
We evaluate the model on held-out sequences from these players and on sequences from 1,000 players entirely unseen during training to assess generalization to unseen identities.
For player identification, a nearest-centroid classifier built from reference embeddings achieves strong performance, reaching balanced accuracies of 0.950 for players seen during training and 0.879 for players unseen during training.
Systematically increasing the number of training players substantially improves generalization to unseen players, indicating a transition from simple fingerprinting to learning transferable behavioral patterns.
\end{abstract}


\newcommand{\meanstd}[2]{#1{\scriptsize$\pm$#2}}

\section{Introduction}\label{sec:intro}

Video games generate large amounts of player input data, capturing complex patterns of player behavior~\cite{sukthankar-profiling-2018}.
While they contain huge amounts of valuable information, important behavioral patterns are mostly hidden behind layers of noise and complexity, making it challenging to extract useful insights without extensive manual and game-specific preprocessing~\cite{game_analytics_book}.
Extracting compact and meaningful representations from these input sequences offers a promising approach that enables a wide range of downstream player behavior analysis tasks, while reducing or eliminating the need for manual analysis~\cite{wang_player2vec_2024}.
Efficiently processing and analyzing large amounts of player input data is highly valuable for game development, as manual analysis is time-consuming, limited in scope, and increasingly impractical for games with many players~\cite{game_analytics_book}.

In this work, we focus on the task of player identification, which is about deciding which individual an input sequence belongs to.
This becomes increasingly challenging as the number of players grows.
Player identification is useful for monitoring account integrity, for example by flagging potential account sharing~\cite{league_account_sharing}, account hijacking~\cite{league_account_hijacking}, and smurfing~\cite{league_smurfing}, where experienced players use fresh novice-rated accounts, undermining matchmaking integrity.

We study player identification using learned representations based on raw input sequences, instead of relying on manual, game-specific feature engineering.
We focus on keystroke and mouse dynamics, which are mostly game-agnostic.
These fine-grained inputs are diverse, but at the same time contain patterns that are consistent per individual, enabling reliable identification~\cite{cont_auth_survey}.
We propose the framework PlayerEmbed, which leverages deep metric learning to extract discriminative and compact embeddings from raw player input sequences, enabling efficient and scalable analysis of player behavior.
By training on a wide range of players with diverse skill levels and playstyles, PlayerEmbed is designed to model the complex patterns and relationships of in-game actions, capturing relevant aspects of temporal dynamics, behavioral style, and skill.
Since embeddings are significantly smaller than the input sequences, they reduce storage requirements and improve computational efficiency in further analysis.

DDraceNetwork (DDNet) is a \enquote{free and open-source side-scrolling platformer with unique cooperative gameplay}~\cite{ddnet_wiki}.
The game's core mechanic centers around using a grappling hook, allowing players to maneuver through complex map layouts with the goal of reaching the finish of a map.
DDNet is particularly well-suited for this work due to several key factors:
The game is free and open-source, allowing self-hosting of game servers and enabling the storage of player input data.
Since the game has existed for many years, there is a large variety in player skill, as the community includes everything from complete beginners to veteran players.
Also, while the game's input schema is relatively simple, the resulting physics-based movement is highly dynamic and complex.
This is crucial as it introduces sufficient diversity in player behavior that makes distinguishing between players non-trivial, as there are no clear repeating discrete patterns.

Our main contributions in this work are as follows.
We introduce PlayerEmbed, a framework that learns compact and discriminative embeddings of player behavior directly from raw input sequences using deep metric learning, without manual feature engineering.
We demonstrate an efficient nearest-centroid player identification method based on PlayerEmbed embeddings, with the encoder trained on up to 1,000 players and evaluated on a further 1,000 players unseen during training, at a scale substantially larger than in prior work.
Our experiments show that the learned representations capture meaningful patterns that generalize even to these new players.

In this work, we start by reviewing related work on player behavior analytics using representation learning in Section~\ref{sec:related_work}.
Next, we introduce the PlayerEmbed framework, describing our dataset, model architecture, and player identification methods in Section~\ref{sec:method}.
We then present and analyze experimental results in Section~\ref{sec:res} and conclude with key findings and directions for future work in Section~\ref{sec:further_work}.

\section{Related Work}\label{sec:related_work}

Research on player behavior in video games has grown increasingly relevant, offering insights into diverse challenges in game development~\cite{sukthankar-profiling-2018}.
Most studies rely on preprocessed aggregated features rather than raw input data~\cite{suznjevic_mmorpg_2011,lee_characterizing_2024,sukthankar-profiling-2018}.
These features are typically derived from gameplay statistics or high-level behavioral summaries, which simplifies analysis but requires substantial game-specific preprocessing.

Several works instead learn representations from game state observations rather than player inputs.
These approaches use images, video, or audio recordings of gameplay~\cite{trivedi_towards_2023,anand_unsupervised_2019,trivedi_contrastive_2021} and have been applied to tasks such as genre recognition~\cite{amiriparian_are_2020} and player experience detection~\cite{makantasis_pixels_2019}.
However, they rely on indirect observations of gameplay instead of the underlying interaction signals produced by the player.
Limited access to raw gameplay data remains a recurring challenge.
This limitation is highlighted by works such as DeepLogger~\cite{intharah_deeplogger_2018}, which attempts to reconstruct player inputs from video recordings.

Some prior work has explored representation learning based on player interaction data, but typically relies on preprocessed formats rather than raw inputs.
Examples include behavior logs with discrete actions~\cite{wang_player2vec_2024,shah_representing_2023}, aggregated statistical gameplay features~\cite{drachen_guns_2012,behaviorformer,factorization_representation}, or provenance graphs describing action dependencies~\cite{melo_player_2020}.
In contrast, representation learning directly from raw game input sequences remains largely unexplored.

Behavioral biometrics based on keystroke dynamics and mouse movements have been widely studied for user authentication.
These interaction signals contain distinctive behavioral patterns that enable identifying individual users~\cite{gamboa_behavioral_2004,zheng_efficient_2016,cont_auth_survey}.
However, many approaches are evaluated on small datasets with fewer than 100 individuals and typically assume closed-set classification where all users must be known during training~\cite{cont_auth_survey}.
Similar problems are present in work on player identification in the video game domain, where recent approaches mostly rely on hand-crafted features and evaluate on small, closed player pools~\cite{league_account_sharing,csgo_identify, beacon}.
In contrast, our work is evaluated on a substantially larger dataset consisting of 1{,}000 known players seen during training and 1{,}000 additional unknown players not included in training.

Deep metric learning provides an effective framework for learning semantically meaningful embeddings~\cite{deep_metric_learn_survey}.
Unlike traditional classification objectives, metric learning produces embeddings that generalize beyond the classes observed during training~\cite{musgrave2020metric}.
Traditional objectives such as Triplet Margin Loss~\cite{triplet_margin_loss} rely on pairwise comparisons between samples but can lead to slow training due to high computational complexity~\cite{Kim_2020_CVPR_proxy}.
Proxy-based approaches address this by introducing representative proxy vectors that reduce the number of comparisons during training.
Proxy Anchor Loss~\cite{Kim_2020_CVPR_proxy} further improves this approach by modeling data-to-data relations while maintaining high training efficiency.

Recent work has also used DDNet gameplay data to predict player churn using aggregated behavioral features together with machine learning and large language models~\cite{ddnet_churn_llm}.
This shows that DDNet gameplay data provides a useful basis for behavioral analytics, although it also relies on aggregated features.

\begin{figure*}
    \centering
    \includegraphics[width=0.84\linewidth]{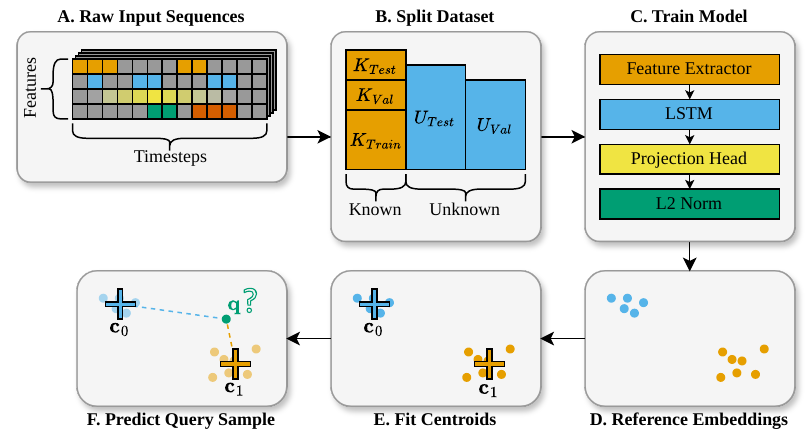}
    \caption{Flowchart of how the PlayerEmbed framework can be used for player identification using nearest-centroid classification.}
    \label{fig:framework}
\end{figure*}

\section{Methodology}\label{sec:method}

\subsection{Data Collection \& Preprocessing}\label{sub:data_collection}

DDNet game servers support optional recording of player inputs.
All player inputs that the game server receives are stored for each update of the DDNet physics engine, which runs at 50 discrete timesteps per second.

As depicted in Step A of Figure~\ref{fig:framework}, PlayerEmbed uses multi-dimensional time sequences as input. 
The framework is agnostic to the number and type of features, supporting both discrete and continuous inputs.
In this work, for each timestep we use an 8-dimensional feature vector comprising the horizontal movement key state encoded as a balanced ternary value (left/none/right), the jump key and two mouse buttons encoded as binary values, and the cursor position in polar form, represented by continuous angle and distance.
To provide some game-state context, we additionally include the player velocity in horizontal and vertical directions.
All continuous features are standardized individually via Z-score normalization using statistics only from training data.

Raw input sequences on 950 different playable maps were collected.
As the input sequences originate from real players, the data were collected anonymously to preserve privacy.
The dataset consists only of raw input sequences associated with randomly assigned player IDs that cannot be traced back to any individual.

The data collection spanned from October 30, 2024, to November 8, 2025.
While many players contribute only limited individual playtime, their combined activity accounts for a substantial portion of the recorded data.
Only a small subset of players provides significant individual playtime.
We restrict the dataset to the 3,000 players with the highest recorded playtime, focusing on players whose available data are sufficient for reliable model training or evaluation.

For each player, the available data are split into sequences with a fixed length of 1,000 timesteps.
In this work, we use sequences of fixed length because they provide a consistent temporal window across all samples and players, while also enabling more efficient data storage and loading due to their uniform shape.
The last timestep is dropped so that delta features such as the player velocity can be calculated for all remaining timesteps. 
Therefore, each sequence still effectively contains the information of 1,000 timesteps, which is equivalent to 20 seconds of player behavior.
For the selected 3,000 players, the number of available sequences ranges from 53 to 2{,}585, corresponding to 17.7 minutes and 14.4 hours of active gameplay, respectively.
As the sequences are unevenly distributed among players, a maximum sample threshold of 1,000 sequences per player is applied.
This corresponds to up to 5.56 hours of active gameplay per player and helps limit the excessive number of sequences from the top 235 players, while still retaining 75\% of the total input sequences.

\subsection{Dataset Construction}\label{sub:dataset}

As depicted in Step B of Figure~\ref{fig:framework}, players are divided into known and unknown sets, with the samples of known players further split into training, validation, and test subsets. 
The samples of unknown players remain unseen during model training.
All players are first sorted by their number of samples in descending order.
The samples of the 1,000 players with the most samples form the known set \(K\).
The available data are split into subsets \(K_{Train}, K_{Val}\) and \(K_{Test}\).
First, a grouped and stratified split holds out approximately 25\% of samples for \(K_{Test}\) by assigning entire play sessions to the test set, while aiming to preserve the proportion of samples per player.
The grouping prevents samples from the same session from appearing in both train and test.
The remaining data are stratified by player ID into train and validation with a 25\% validation split, resulting in overall proportions of approximately 56\% \(K_{Train}\), 19\% \(K_{Val}\), and 25\% \(K_{Test}\).

The next 1,000 players form the unknown test set \(U_{Test}\), and the subsequent 1,000 players form the unknown validation set \(U_{Val}\).
We assign players with the highest number of samples to \(K\), as their samples are further divided into subsets, and players with the next-highest number of samples to \(U_{Test}\), to enable a robust final evaluation on both test sets.
\(K_{Train}\) is used to train the model, while \(K_{Val}\) and \(U_{Val}\) are used to monitor performance on known and unknown players during training.
After training, \(K_{Test}\) evaluates held-out samples of known players, while \(U_{Test}\) evaluates unknown players unseen during model training.

\subsection{Model}\label{model}

In this section, we present a hybrid CNN-LSTM architecture to create compact and meaningful representations from raw player input sequences.
Step C in Figure~\ref{fig:framework} provides an overview of the model. 

In-game actions are typically produced by combinations of inputs within short time windows. We therefore first use convolutional layers for efficient local pattern extraction, so that the LSTM can focus on modeling long-range temporal dependencies. 
The CNN feature extractor is inspired by MobileNetV1~\cite{mobilenetsV1}, using depthwise separable convolutions, which factorize a standard convolution into a depthwise convolution and a 1$\times$1 pointwise convolution.
This significantly reduces computational cost compared to standard convolutions.
The depthwise convolution extracts short per-feature patterns within its kernel window, while the following pointwise convolution mixes information across features, capturing short cross-feature interactions.
The model applies a 1D depthwise convolution with a kernel size of five and no padding, using 16 depthwise filters per input feature.
With eight input features, this yields 128 channels per timestep.
These channels are processed by batch normalization and ReLU, followed by a pointwise convolution with 128 output channels, again followed by batch normalization and ReLU.
We add channel-wise dropout with probability $p=0.1$, zeroing out feature channels to reduce over-reliance on individual features.

The extracted features are passed into a unidirectional single-layer LSTM with hidden size 128.
As the LSTM iteratively processes timesteps, its hidden state accumulates information from previous timesteps.
The hidden state of the last timestep effectively summarizes the entire sequence without further pooling and is used as the basis for the final embedding.
This final hidden state vector is batch normalized and fed into a two-layer MLP projection head with hidden dimension 256 and a GELU activation, followed by L2 normalization to obtain the final 128-dimensional output embeddings.

Architecture and hyperparameters were selected based on preliminary experiments on a separate, smaller dataset collected on different game servers without any overlap with the data used in this work.
We did not further tune the architecture on the final dataset to avoid bias.
Additional tuning on the validation subsets might further improve performance.

\subsection{Training}\label{training}

We train the model with batches of 96 sequences and a base learning rate of $10^{-3}$.
For each batch, we compute the model’s L2-normalized embeddings and optimize a Proxy Anchor Loss~\cite{Kim_2020_CVPR_proxy} with default parameters (margin $\delta = 0.1$, scaling factor $\alpha = 32$) using cosine similarity as the similarity measure between embeddings.
We chose Proxy Anchor Loss as it led to substantially faster convergence than traditional metric learning objectives, such as Triplet Margin Loss~\cite{triplet_margin_loss}, reducing training time by several hours for larger numbers of players.
We use random uniform sampling over sequences.
In preliminary experiments, class-weighted sampling and MPerClass batching did not measurably improve performance over uniform sampling, so we use the simpler method.

Model weights and learnable proxies are updated using separate Adam optimizers with default parameters.
To prevent exploding gradients in the LSTM component, we apply gradient clipping with a maximum norm of 10.
Learning rate reduction by a factor of 0.8 is utilized if the validation loss (on \(K_{Val}\)) does not improve by at least 1\% over 10 consecutive epochs, to improve model convergence during later stages of training.
Training is halted if the validation loss does not improve by at least 1\% over 20 consecutive epochs.
The final model is selected based on the lowest observed validation loss and then used for the final evaluation on the held-out test sets \(K_{Test}\) and \(U_{Test}\).

\subsection{Player Identification}

We evaluate the quality of the learned embeddings for player identification using similarity-based classification methods.
Let $\mathbf{x} \in \mathbb{R}^d$ denote an embedding, and define cosine similarity as $s(\mathbf{a}, \mathbf{b}) = \mathbf{a}^\top \mathbf{b} / (\|\mathbf{a}\|_2 \|\mathbf{b}\|_2)$.
Figure~\ref{fig:framework} (Steps D--F) illustrates a simplified example with two reference classes, demonstrated using the nearest-centroid method.
In this work, we compare nearest-centroid, nearest-medoid, and k-nearest neighbors (k-NN).

For nearest-centroid classification, we first collect a set of reference embeddings $R_p = \{\mathbf{x}_{p,i}\}_{i=1}^{N_p}$ for each player ID $p$ (Step D). Prototype centroids are computed as
$\tilde{\mathbf{c}}_p = \frac{1}{N_p} \sum_i \mathbf{x}_{p,i}$ and then re-normalized $\mathbf{c}_p = \tilde{\mathbf{c}}_p / \|\tilde{\mathbf{c}}_p\|_2$ (Step E). During inference (Step F), the predicted player ID for a query embedding $\mathbf{q}$ is
$\hat{y}(\mathbf{q}) = \arg\max_p s(\mathbf{c}_p, \mathbf{q})$.

Nearest-medoid classification follows the same prediction rule, but prototypes are defined as
$\mathbf{m}_p = \arg\max_{\mathbf{x} \in R_p} \frac{1}{N_p} \sum_i s(\mathbf{x}, \mathbf{x}_{p,i})$,
i.e., the reference embedding with the highest average cosine similarity to other reference embeddings of the same player. This approach avoids averaging and can be more robust to outliers.

For k-NN, prediction is based on a majority vote among the $k$ reference embeddings most similar to the query. Let $\mathcal{N}_k(\mathbf{q})$ be the indices of the $k$ nearest neighbors with labels $y_i$. 
Then we predict $\hat{y}(\mathbf{q}) = \arg\max_c \left|\{ i \in \mathcal{N}_k(\mathbf{q}) : y_i = c \}\right|$.

To improve robustness, we average $M$ query embeddings $\{\mathbf{q}_j\}_{j=1}^M$, re-normalize the result, and use this aggregated vector
$\bar{\mathbf{q}} = \big(\sum_j \mathbf{q}_j\big) \big/ \big\|\sum_j \mathbf{q}_j\big\|_2$
as the query to obtain a single prediction for all $M$ samples. As this multi-sample averaging reduces the number of samples available for evaluation, we only consider small $M$. We evaluate $M \in \{1,3,6\}$, where $M=1$ uses one embedding per 20-second sequence, $M=3$ summarizes one minute, and $M=6$ summarizes two minutes in a single embedding.

To evaluate a classification method, we split the samples of each player into disjoint reference and query sets using the same grouped and stratified procedure as in Section~\ref{sub:dataset}, ensuring a balanced split without leakage across play sessions.
We use a 50/50 split to give equal importance to reference and query embeddings, ensuring there are enough samples for both fitting and evaluation.
We repeat this procedure with five random seeds, reporting averaged metrics to reduce variance due to specific data splits.
Reference embeddings are available for both known and unknown players, so the unknown setting measures transfer of the embedding model to unseen identities.

We aim to evaluate classification performance on known and unknown players, both individually and jointly, while also capturing potential collisions between their embeddings.
To indicate which reference and query sets are used in each evaluation, we introduce the notation \(R \to Q\), where \(R\) indicates the reference subset(s) used to construct the classifier and \(Q\) the query subset(s) used for evaluation.
For example, \(\text{F1}(KU \to U)\) denotes the F1-score obtained when using the reference sets of both \(K\) and \(U\), while evaluating predictions only on the query samples from \(U\).
Rather than evaluating unknown players in isolation, this measures their performance while explicitly considering collisions with known players.
Unless stated otherwise, metrics are reported on the corresponding test splits \(K_{Test}\) and \(U_{Test}\) (abbreviated as \(K\) and \(U\) for readability).

\section{Experiments}\label{sec:res}

\subsection{Experimental Setup}

We conduct experiments that vary the number of known players used for training.
Specifically, we sweep over \(n \in \{100, \dots, 1000\}\), which restricts the known set \(K\) to the first \(n\) players (sorted by sample count) before splitting into \(K_{Train}\), \(K_{Val}\) and \(K_{Test}\).
The unknown sets \(U_{Test}\) and \(U_{Val}\) remain fixed across all experiments.
For each \(n\), we repeat the experiment with five random seeds and report the aggregated results as mean $\pm$ std, reducing noise introduced by specific data splits or non-deterministic model training.
All models are trained using NVIDIA Tesla V100 GPUs.

Using the test embeddings, we analyze different classification methods (Section~\ref{sub:classification_methods}) and $n$ values (Section~\ref{sub:n_values}).
Then, to contextualize our findings, we compare PlayerEmbed to traditional baselines based on hand-crafted statistical features (Section~\ref{sub:traditional_baseline}) and perform an ablation study for our proposed architecture including alternative encoder models (Section~\ref{sub:ablation}).
Finally, we analyze the embedding space using dimensionality reduction and clustering (Section~\ref{sub:embedding_analysis}).

Because the class distribution is imbalanced, we compute all evaluation metrics in a macro-averaged fashion to give equal importance to each player.
We focus on balanced accuracy to fairly evaluate the model across different numbers of players, as it provides a straightforward measure of per-player performance.

\subsection{Classification Method Comparison}\label{sub:classification_methods}

\begin{table}[t]
\caption{Balanced accuracy of classification methods across evaluation configs and averaging levels $M$ for $n=1000$.}
\label{tab:all_results}
\centering
\setlength{\tabcolsep}{2.6pt}
\renewcommand{\arraystretch}{1.05}
\begin{tabular}{@{}llccccc@{}}
\toprule
Config & $M$ & Centroid & Medoid & 1-NN & 5-NN & 9-NN \\
\midrule
\rowcolor{gray!10}
K$\to $K & 1 &
\textbf{\meanstd{0.790}{.007}} & \meanstd{0.720}{.010} & \meanstd{0.742}{.008} & \meanstd{0.743}{.008} & \meanstd{0.740}{.008} \\
K$\to$K & 3 &
\textbf{\meanstd{0.931}{.004}} & \meanstd{0.889}{.005} & \meanstd{0.889}{.004} & \meanstd{0.886}{.004} & \meanstd{0.881}{.004} \\
\rowcolor{gray!10}
K$\to$K & 6 &
\textbf{\meanstd{0.950}{.003}} & \meanstd{0.920}{.004} & \meanstd{0.917}{.004} & \meanstd{0.913}{.004} & \meanstd{0.908}{.004} \\
\midrule
KU$\to$U & 1 &
\textbf{\meanstd{0.621}{.008}} & \meanstd{0.520}{.008} & \meanstd{0.599}{.007} & \meanstd{0.598}{.007} & \meanstd{0.601}{.007} \\
\rowcolor{gray!10}
KU$\to$U & 3 &
\textbf{\meanstd{0.833}{.004}} & \meanstd{0.728}{.006} & \meanstd{0.775}{.005} & \meanstd{0.782}{.005} & \meanstd{0.783}{.004} \\
KU$\to$U & 6 &
\textbf{\meanstd{0.879}{.004}} & \meanstd{0.792}{.007} & \meanstd{0.827}{.005} & \meanstd{0.832}{.005} & \meanstd{0.831}{.005} \\
\bottomrule
\end{tabular}
\end{table}

Table~\ref{tab:all_results} compares different similarity-based classification methods with balanced accuracy for both known and unknown player evaluation scenarios.
We report results for different levels of multi-sample averaging $M \in \{1, 3, 6\}$.

Across different values of $k$, k-NN exhibits marginal performance differences, suggesting that local neighborhoods in the embedding space are relatively homogeneous.
However, we observe a trend where increasing $k$ leads to slightly worse performance on known players, but small improvements for unknown players, indicating that embeddings for unknown players are more noisy and benefit from majority voting.

Centroid-based classification consistently achieves the best performance.
It outperforms medoid and k-NN for both known and unknown player configurations and for all $M$.
This indicates that the embedding space benefits from mean pooling as the Proxy Anchor Loss pulls each player's embeddings into a single compact cluster, while k-NN remains sensitive to individual noisy reference embeddings and medoid prototypes do not benefit from averaging.

Comparing the runtime, both centroid- and medoid-based classification complete in under two seconds across all configurations, while k-NN scales linearly with the number of reference samples and ranges from 10.4 to 115.3 seconds depending on $M$ and the reference sets, with only marginal impact by $k$.
This large disparity highlights the practical limitations of k-NN, while demonstrating that centroid and medoid methods are suited for practical use in terms of runtime.

\subsection{Scaling of Player Identification}\label{sub:n_values}

\begin{table}
\caption{Balanced accuracy and macro F1 for nearest-centroid classification across evaluation configurations, numbers of known training players $n$, and averaging levels $M$.}

\label{tab:centroid_performance}
\centering
\small
\setlength{\tabcolsep}{3pt}
\renewcommand{\arraystretch}{1.05}
\begin{tabular}{@{}llcccc@{}}
\toprule
& & \multicolumn{2}{c}{Balanced Accuracy} & \multicolumn{2}{c}{Macro F1} \\
\cmidrule(lr){3-4}\cmidrule(l){5-6}
Config & $n$ & $M=1$ & $M=6$ & $M=1$ & $M=6$ \\
\midrule
\rowcolor{gray!10}
K$\to$K & 100 &
\meanstd{0.8990}{.009} & \meanstd{0.9998}{.000} &
\meanstd{0.8985}{.009} & \meanstd{0.9998}{.000} \\

K$\to$K & 200 &
\meanstd{0.8993}{.004} & \meanstd{0.9996}{.000} &
\meanstd{0.8977}{.004} & \meanstd{0.9996}{.000} \\

\rowcolor{gray!10}
K$\to$K & 400 &
\meanstd{0.8900}{.006} & \meanstd{0.9967}{.001} &
\meanstd{0.8825}{.006} & \meanstd{0.9964}{.001} \\

K$\to$K & 600 &
\meanstd{0.8564}{.007} & \meanstd{0.9851}{.004} &
\meanstd{0.8407}{.008} & \meanstd{0.9826}{.005} \\

\rowcolor{gray!10}
K$\to$K & 800 &
\meanstd{0.8281}{.009} & \meanstd{0.9704}{.005} &
\meanstd{0.8015}{.009} & \meanstd{0.9641}{.006} \\

K$\to$K & 1000 &
\meanstd{0.7899}{.007} & \meanstd{0.9503}{.003} &
\meanstd{0.7532}{.007} & \meanstd{0.9385}{.005} \\

\midrule
\rowcolor{gray!10}
KU$\to$U & 100 &
\meanstd{0.2217}{.015} & \meanstd{0.6359}{.016} &
\meanstd{0.2168}{.014} & \meanstd{0.5975}{.017} \\

KU$\to$U & 200 &
\meanstd{0.2803}{.021} & \meanstd{0.7134}{.011} &
\meanstd{0.2880}{.020} & \meanstd{0.6806}{.012} \\

\rowcolor{gray!10}
KU$\to$U & 400 &
\meanstd{0.4711}{.017} & \meanstd{0.8227}{.005} &
\meanstd{0.4794}{.015} & \meanstd{0.7994}{.005} \\

KU$\to$U & 600 &
\meanstd{0.5559}{.007} & \meanstd{0.8556}{.002} &
\meanstd{0.5646}{.006} & \meanstd{0.8382}{.003} \\

\rowcolor{gray!10}
KU$\to$U & 800 &
\meanstd{0.5997}{.010} & \meanstd{0.8729}{.004} &
\meanstd{0.6112}{.010} & \meanstd{0.8591}{.004} \\

KU$\to$U & 1000 &
\meanstd{0.6210}{.008} & \meanstd{0.8786}{.004} &
\meanstd{0.6365}{.008} & \meanstd{0.8685}{.005} \\
\bottomrule
\end{tabular}
\end{table}

Table~\ref{tab:centroid_performance} reports balanced accuracy and macro F1-score for the proposed nearest-centroid classification method for increasing numbers of known players \(n \in \{100,\dots,1000\}\) and multi-sample averaging with \(M \in \{1,6\}\).
For brevity, results for the intermediate setting \(M=3\) are not shown.
They consistently fall between those for \(M=1\) and \(M=6\), generally closer to \(M=6\), indicating a smooth performance improvement as more sequences are averaged.
Two evaluation configurations are used to analyze the individual performance on known and unknown players: \(K \to K\) evaluates identification of known players only, and \(KU \to U\) evaluates unknown players while accounting for collisions with known-player centroids.

Across all configurations, standard deviations are consistently low, indicating stable model training and reliable evaluation across different data splits.
Balanced accuracy and macro F1-score closely match each other.
The similarity of balanced accuracy and F1 suggests that the model maintains a good balance between recall and precision across players.

For known players, performance is highest when the number of known players $n$ is small.
For \(n=100\), a balanced accuracy of \(0.8990\) is achieved with single-sample predictions ($M=1$).
By using multi-sample averaging ($M>1$), near-perfect identification performance is obtained.
As \(n\) increases, performance on known players gradually declines, reflecting the increasing difficulty of separating a larger number of players in the embedding space.
Nevertheless, even for \(n=1000\), strong performance is maintained, reaching a balanced accuracy of \(0.7899\) for \(M=1\) and \(0.9503\) with \(M=6\).
These results demonstrate good generalization for known players, enabling strong player identification scaling up to 1,000 known players.

In contrast, performance for unknown players is very limited when the model is trained with a small number of known players \(n\), only reaching a balanced accuracy of \(0.2217\) for \(n=100\) with \(M=1\).
Reasonable identification performance emerges once \(n\) reaches higher values, increasing to \(0.6210\) for \(n=1000\) and \(M=1\).
Multi-sample averaging further improves performance, raising the balanced accuracy to \(0.8786\) with \(M=6\).
These results demonstrate that, by increasing $n$, good generalization can also be achieved for unknown players, allowing reliable player identification even for 1,000 unseen players.

We observe that generalization to known and unknown players does not necessarily align.
Decent generalization to unknown players emerges once the number of known players is increased.
This emergent behavior suggests that, as the number of known players grows, the model is forced to learn representations that generalize beyond known players and capture more fundamental and transferable aspects of player behavior.
This shift aligns with the systematic decrease in performance observed for known players, indicating a trade-off between specialization and generalization in the learned embedding space.

\subsection{Traditional Baseline}\label{sub:traditional_baseline}

To contextualize the achieved player identification performance using learned embeddings, we introduce a traditional baseline that follows common practice in game analytics and behavioral biometrics, aggregating sequential data into fixed-size representations using hand-crafted statistical features~\cite{drachen_guns_2012,game_analytics_book,gamboa_behavioral_2004,zheng_efficient_2016,cont_auth_survey,league_account_sharing}.
For discrete inputs, we collect state frequencies, transition rates, and state durations. 
For continuous inputs, we use raw value statistics, temporal change dynamics, and histogram-based properties. 
In addition, we compute autocorrelations with different lag intervals for all features. 
The resulting 151-dimensional feature vectors are used both with traditional machine learning classifiers (MLP, SVM, XGBoost) and with nearest-centroid classification.
For a fair comparison, we tune the hyperparameters of the traditional classifiers and also consider multi-sample averaging.

\begin{table}
\caption{Balanced accuracy of traditional baselines compared to \mbox{PlayerEmbed}. Bold marks the best baseline.}

\label{tab:statfeat_comparison}
\centering
\small
\setlength{\tabcolsep}{2.0pt}
\renewcommand{\arraystretch}{1.05}
\begin{tabular}{@{}lcccc@{}}
\toprule
& \multicolumn{2}{c}{$n=100$} & \multicolumn{2}{c}{$n=1000$} \\
\cmidrule(lr){2-3} \cmidrule(l){4-5}
Method & $M=1$ & $M=6$ & $M=1$ & $M=6$ \\
\midrule
\rowcolor{gray!10}
MLP &
\textbf{\meanstd{0.5522}{.0060}} & \meanstd{0.7633}{.0068} &
\textbf{\meanstd{0.2513}{.0015}} & \textbf{\meanstd{0.3946}{.0082}} \\

SVM &
\meanstd{0.4407}{.0024} & \textbf{\meanstd{0.7951}{.0077}} &
\meanstd{0.1332}{.0013} & \meanstd{0.3549}{.0042} \\

\rowcolor{gray!10}
Centroid &
\meanstd{0.3197}{.0040} & \meanstd{0.7566}{.0072} &
\meanstd{0.1226}{.0006} & \meanstd{0.3701}{.0056} \\

XGBoost &
\meanstd{0.5418}{.0049} & \meanstd{0.7124}{.0055} &
\meanstd{0.2200}{.0026} & \meanstd{0.3148}{.0053} \\

\midrule
\rowcolor{gray!10}
PlayerEmbed &
\meanstd{0.8990}{.0090} & \meanstd{0.9998}{.0000} &
\meanstd{0.7899}{.0070} & \meanstd{0.9503}{.0030} \\
\bottomrule
\end{tabular}
\end{table}

Table~\ref{tab:statfeat_comparison} compares the traditional baselines to PlayerEmbed for different numbers of known players \(n\) and multi-sample averaging levels \(M\).
All traditional classifiers benefit noticeably from multi-sample averaging with \(M=6\), which consistently boosts balanced accuracy. 
SVM and nearest-centroid classification in particular require \(M=6\) to reach competitive performance, while MLP performs best among the traditional methods in most configurations.
For \(n=100\), SVM achieves the highest balanced accuracy of \(0.7951\) with \(M=6\), slightly outperforming MLP.
However, when increasing the number of known players to \(n=1000\), performance drops substantially for all traditional methods.
Even with \(M=6\), the best balanced accuracy is only \(0.3946\).
In contrast, PlayerEmbed maintains strong performance across all settings and scales much better with the number of players.
While the traditional, hand-crafted feature representations yield non-trivial identification performance, especially for small player counts, PlayerEmbed substantially outperforms them, particularly in the more challenging large-scale setting.
These findings highlight both the increased difficulty of separating many players and the strong scalability and effectiveness of learned PlayerEmbed representations.

\subsection{Encoder Architecture Evaluation}\label{sub:ablation}

We perform experiments to measure the impact of the employed encoder architecture in our PlayerEmbed framework.
Specifically, we replace the CNN-LSTM architecture with a Transformer encoder~\cite{transformers} and an MLP.
The MLP operates on flattened input, using hidden sizes [512, 256] with GELU activations and dropout 0.5.
The Transformer encoder uses a hidden dimension of 128, mean pooling, and dropout 0.1, with a variable number of layers $L$.
We also evaluate ablations of our CNN-LSTM architecture: removing the CNN feature extractor, denoted by the missing \textit{CNN-} prefix, and removing the projection head, denoted by the \textit{(no head)} suffix.
When the projection head is disabled, the batch-normalized final LSTM hidden state is directly L2-normalized and used as output.
We train with $n=1000$ and reuse the same five random seeds as in the previous experiment.

\begin{table}
\caption{Balanced accuracy for $n=1000$ on known and unknown players across averaging levels $M$. Rows are ordered by known-player performance at $M=6$.}
\label{tab:model_comparison}
\centering
\small
\setlength{\tabcolsep}{2.25pt}
\renewcommand{\arraystretch}{1.05}
\begin{tabular}{@{}lcccc@{}}
\toprule
Model & \multicolumn{2}{c}{K$\to$K} & \multicolumn{2}{c}{KU$\to$U} \\
\cmidrule(lr){2-3} \cmidrule(l){4-5}
& $M=1$ & $M=6$ & $M=1$ & $M=6$ \\
\midrule
\rowcolor{gray!10}
CNN-LSTM &
\textbf{\meanstd{0.790}{.007}} & \textbf{\meanstd{0.950}{.003}} &
\textbf{\meanstd{0.621}{.008}} & \textbf{\meanstd{0.879}{.004}} \\
Transformer$_{L=3}$ &
\meanstd{0.698}{.005} & \meanstd{0.931}{.002} &
\meanstd{0.385}{.010} & \meanstd{0.785}{.006} \\
\rowcolor{gray!10}
CNN-LSTM$_{\text{no head}}$ &
\meanstd{0.702}{.017} & \meanstd{0.921}{.009} &
\meanstd{0.524}{.018} & \meanstd{0.846}{.009} \\
Transformer$_{L=2}$ &
\meanstd{0.681}{.004} & \meanstd{0.921}{.002} &
\meanstd{0.444}{.009} & \meanstd{0.803}{.003} \\
\rowcolor{gray!10}
LSTM &
\meanstd{0.680}{.027} & \meanstd{0.914}{.009} &
\meanstd{0.506}{.020} & \meanstd{0.830}{.012} \\
LSTM$_{\text{no head}}$ &
\meanstd{0.546}{.031} & \meanstd{0.851}{.018} &
\meanstd{0.403}{.022} & \meanstd{0.765}{.017} \\
\rowcolor{gray!10}
Transformer$_{L=1}$ &
\meanstd{0.415}{.037} & \meanstd{0.783}{.030} &
\meanstd{0.304}{.020} & \meanstd{0.692}{.021} \\
MLP &
\meanstd{0.068}{.002} & \meanstd{0.243}{.005} &
\meanstd{0.051}{.001} & \meanstd{0.206}{.003} \\
\bottomrule
\end{tabular}
\end{table}

Table~\ref{tab:model_comparison} reports the impact on balanced accuracy of different architectures for player identification on known and unknown players.
Removing either the CNN feature extraction or the projection head substantially reduces performance, especially for $M=1$.
When disabling both, this becomes even more pronounced, with a relative balanced accuracy decrease of $30.9\%$ for known and $35.1\%$ for unknown players at $M=1$, and $10.4\%$ and $12.9\%$ at $M=6$.
This shows that both components are crucial for strong generalization to known and unknown players.
The Transformer architecture provides a strong baseline, achieving good performance on known players as the number of layers increases.
With $L=3$ layers and $M=6$, it reaches a balanced accuracy of $0.931$ on known players, almost matching CNN-LSTM.
However, this comes at the cost of reduced generalization to unknown players, where the best unknown performance of $0.803$ is obtained with $L=2$.
Moreover, Transformer performance benefits more from multi-sample averaging, which suggests overall noisier embeddings.
Also, Transformer training is significantly slower.
For $n=1000$, CNN-LSTM trains in $4.5$ hours on average, whereas Transformer requires $8.1$ hours ($L=1$), $22.1$ hours ($L=2$), and $28.6$ hours ($L=3$), while still achieving worse overall performance.
MLP does not reach practical identification performance, performing far worse than both Transformer and CNN-LSTM.
This highlights that an architecture with sequential modeling capabilities is crucial to extract good representations.

Overall, our proposed CNN-LSTM architecture achieves the best performance for known and unknown players while maintaining substantially lower training time than Transformer.

\subsection{Embedding Space Analysis}\label{sub:embedding_analysis}

We perform a qualitative analysis of the resulting embedding space using dimensionality reduction and clustering. 
All embeddings for \(K_{Test}\) and \(U_{Test}\) are collected to compute a representative centroid for each player.
Then we perform dimensionality reduction to two dimensions with UMAP.
In this embedding space we apply the HDBSCAN clustering algorithm to extract players that end up in distinct clusters.

\begin{figure}
    \centering
    \includegraphics[width=\linewidth]{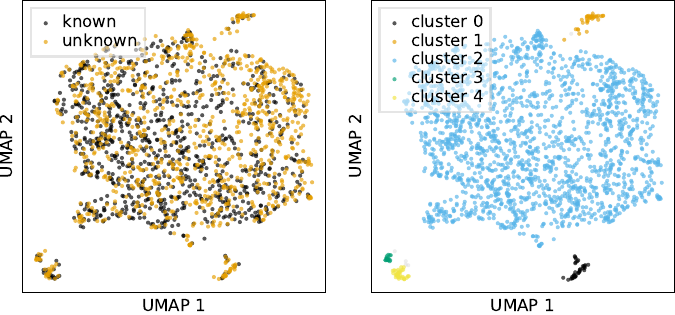}
    \caption{
    UMAP visualization of player centroids for $K_{Test}$ and $U_{Test}$ with HDBSCAN cluster assignments.}
    \label{fig:embedding_analysis}
\end{figure}

Figure~\ref{fig:embedding_analysis} visualizes the dimensionality reduction and the resulting cluster assignments.
The majority of players are assigned to one large, contiguous cluster. 
While some local concentrations of players exist, the embedding space overall appears smooth and mostly uniform, and includes both known and unknown players in overlapping regions.

In addition, four clearly separated, smaller, and dense clusters emerge, each containing 20 to 30 players. 
We manually inspected the gameplay of the players and found that each of the four clusters corresponds to one clear game client modification. 
Players in clusters 1 and 4 use non-default client configurations that consistently alter some of the input features. 
More notably, manual inspection of clusters 0 and 3 revealed unusual, non-human input behavior.
Cluster 0 exhibits input precision and speed far beyond human capability, while cluster 3 shows cursor movements that snap to positions, indicating aim assistance.
Both point to the use of cheat clients, and their behavior differs so strongly from that of regular players that they form two distinct clusters.

These findings indicate that PlayerEmbed learns a smooth and meaningful embedding space that is sensitive to systematic changes in client behavior and goes beyond just learning player identities by heart.

\section{Conclusion and Further Work}\label{sec:further_work}

This work introduces the PlayerEmbed framework to learn compact embeddings from raw game input sequences to support player behavior analysis, with a primary focus on player identification.
PlayerEmbed achieves strong player identification on 1,000 known players and 1,000 additional players that are not part of model training.
In comparison, classifiers based on hand-crafted statistical features achieve moderate performance for small numbers of players but degrade as the number of players grows.
We find that using larger numbers of known training players systematically improves generalization to unseen players and that averaging embeddings over multiple sequences consistently improves performance across all evaluation settings.
Our experiments show that all components of the CNN-LSTM encoder contribute to its performance, and that the full model outperforms LSTM-only, Transformer, and MLP baselines while training substantially faster than the Transformer variants.
Analysis of player prototypes in the embedding space reveals a mostly smooth shared structure for known and unknown players, together with small, well-separated clusters corresponding to specific client modifications and cheat clients.

In conclusion, our results demonstrate that PlayerEmbed encodes player-specific temporal patterns of raw input sequences into compact embeddings that enable robust player identification for both known and unknown players.
The transfer to unseen identities and the observed cluster structure suggest that the embeddings capture meaningful behavioral patterns beyond just player identity.
This indicates that the framework might be useful for other downstream tasks.

Further work could scale PlayerEmbed beyond the current 1,000 known and 1,000 unknown players by collecting larger and more diverse datasets across different games.
No component of PlayerEmbed is tailored specifically to the input features used in this work, and since these rely primarily on mouse dynamics and keystrokes, which are present in most games, we expect the framework to transfer to other games where raw input sequences are available.

\clearpage

\bibliographystyle{IEEEtran}
\bibliography{references}

@inbook{sukthankar-profiling-2018, place={Cambridge}, title={Profiling in Games: Understanding Behavior from Telemetry}, booktitle={Social Interactions in Virtual Worlds: An Interdisciplinary Perspective}, publisher={Cambridge University Press}, author={Sifa, Rafet and Drachen, Anders and Bauckhage, Christian}, year={2018}, pages={337–374}}

@inproceedings{suznjevic_mmorpg_2011,
author = {Suznjevic, Mirko and Stupar, Ivana and Matijasevic, Maja},
title = {MMORPG player behavior model based on player action categories},
year = {2011},
isbn = {9781457719349},
publisher = {IEEE Press},
booktitle = {Proceedings of the 10th Annual Workshop on Network and Systems Support for Games},
articleno = {6},
numpages = {6},
location = {Ottawa, Canada},
series = {NetGames '11}
}

@inproceedings{lee_characterizing_2024,
author = {Lee, Hanbyeol and Lee, Seyeon and Nallapati, Rohan and Uh, Youngjung and Lee, Byungjoo},
title = {Characterizing and Quantifying Expert Input Behavior in League of Legends},
year = {2024},
isbn = {9798400703300},
publisher = {Association for Computing Machinery},
address = {New York, NY, USA},
url = {https://doi.org/10.1145/3613904.3642588},
doi = {10.1145/3613904.3642588},
booktitle = {Proceedings of the 2024 CHI Conference on Human Factors in Computing Systems},
articleno = {957},
numpages = {21},
location = {Honolulu, HI, USA},
series = {CHI '24}
}

@article{trivedi_towards_2023,
  title={Towards general game representations: decomposing games pixels into content and style},
  author={Trivedi, Chintan and Makantasis, Konstantinos and Liapis, Antonios and Yannakakis, Georgios N},
  journal={ArXiv},
  year={2023}
}

@inbook{anand_unsupervised_2019,
author = {Anand, Ankesh and Racah, Evan and Ozair, Sherjil and Bengio, Yoshua and C\^{o}t\'{e}, Marc-Alexandre and Hjelm, R. Devon},
title = {Unsupervised state representation learning in atari},
year = {2019},
publisher = {Curran Associates Inc.},
address = {Red Hook, NY, USA},
booktitle = {Proceedings of the 33rd International Conference on Neural Information Processing Systems},
articleno = {787},
numpages = {14}
}

@ARTICLE{amiriparian_are_2020,
  author={Amiriparian, Shahin and Cummins, Nicholas and Gerczuk, Maurice and Pugachevskiy, Sergey and Ottl, Sandra and Schuller, Björn},
  journal={IEEE Transactions on Games}, 
  title={“Are You Playing a Shooter Again?!” Deep Representation Learning for Audio-Based Video Game Genre Recognition}, 
  year={2020},
  volume={12},
  number={2},
  pages={145-154},
  doi={10.1109/TG.2019.2894532}}

@INPROCEEDINGS{makantasis_pixels_2019,
  author={Makantasis, Konstantinos and Liapis, Antonios and Yannakakis, Georgios N.},
  booktitle={2019 8th International Conference on Affective Computing and Intelligent Interaction (ACII)}, 
  title={From Pixels to Affect: A Study on Games and Player Experience}, 
  year={2019},
  volume={},
  number={},
  pages={1-7},
  doi={10.1109/ACII.2019.8925493}}

@inproceedings{intharah_deeplogger_2018,
author = {Intharah, Thanapong and Brostow, Gabriel J.},
title = {DeepLogger: Extracting User Input Logs From 2D Gameplay Videos},
year = {2018},
isbn = {9781450356244},
publisher = {Association for Computing Machinery},
address = {New York, NY, USA},
url = {https://doi.org/10.1145/3242671.3242674},
doi = {10.1145/3242671.3242674},
booktitle = {Proceedings of the 2018 Annual Symposium on Computer-Human Interaction in Play},
pages = {221–230},
numpages = {10},
location = {Melbourne, VIC, Australia},
series = {CHI PLAY '18}
}

@article{wang_player2vec_2024,
  title={player2vec: A language modeling approach to understand player behavior in games},
  author={Wang, Tianze and Honari-Jahromi, Maryam and Katsarou, Styliani and Mikheeva, Olga and Panagiotakopoulos, Theodoros and Asadi, Sahar and Smirnov, Oleg},
  journal={ArXiv},
  year={2024}
}

@inproceedings{shah_representing_2023,
  author={Shah, Jinay and Thue, David},
  booktitle={2023 IEEE Conference on Games (CoG)}, 
  title={Representing Player Behaviour via Graph Embedding Techniques: A Case Study in Dota 2}, 
  year={2023},
  pages={1-8},
  doi={10.1109/CoG57401.2023.10333216}}

@INPROCEEDINGS{drachen_guns_2012,
  author={Drachen, Anders and Sifa, Rafet and Bauckhage, Christian and Thurau, Christian},
  booktitle={2012 IEEE Conference on Computational Intelligence and Games (CIG)}, 
  title={Guns, swords and data: Clustering of player behavior in computer games in the wild}, 
  year={2012},
  volume={},
  number={},
  pages={163-170},
  doi={10.1109/CIG.2012.6374152}}

@inproceedings{melo_player_2020,
author = {Melo, Sidney Araujo and Kohwalter, Troy C. and Clua, Esteban and Paes, Aline and Murta, Leonardo},
title = {Player Behavior Profiling through Provenance Graphs and Representation Learning},
year = {2020},
isbn = {9781450388078},
publisher = {Association for Computing Machinery},
address = {New York, NY, USA},
url = {https://doi.org/10.1145/3402942.3402961},
doi = {10.1145/3402942.3402961},
booktitle = {Proceedings of the 15th International Conference on the Foundations of Digital Games},
articleno = {62},
numpages = {11},
location = {Bugibba, Malta},
series = {FDG '20}
}

@inproceedings{gamboa_behavioral_2004,
author = {Hugo Gamboa and Ana Fred},
title = {{A behavioral biometric system based on human-computer interaction}},
volume = {5404},
booktitle = {Biometric Technology for Human Identification},
editor = {Anil K. Jain and Nalini K. Ratha},
organization = {International Society for Optics and Photonics},
publisher = {SPIE},
pages = {381 -- 392},
year = {2004},
doi = {10.1117/12.542625},
URL = {https://doi.org/10.1117/12.542625}
}

@article{zheng_efficient_2016,
author = {Zheng, Nan and Paloski, Aaron and Wang, Haining},
title = {An Efficient User Verification System Using Angle-Based Mouse Movement Biometrics},
year = {2016},
issue_date = {April 2016},
publisher = {Association for Computing Machinery},
address = {New York, NY, USA},
volume = {18},
number = {3},
issn = {1094-9224},
url = {https://doi.org/10.1145/2893185},
doi = {10.1145/2893185},
journal = {ACM Trans. Inf. Syst. Secur.},
month = apr,
articleno = {11},
numpages = {27}
}

@INPROCEEDINGS{ddnet_churn_llm,
  author={Schneider, Tobias and Sparrenberg, Lorenz and Sifa, Rafet},
  booktitle={2025 IEEE 12th International Conference on Data Science and Advanced Analytics (DSAA)}, 
  title={{Predicting Player Churn with LLMs: A Comprehensive Evaluation of World Knowledge and Reasoning}}, 
  year={2025},
  volume={},
  number={},
  pages={1-10},
  doi={10.1109/DSAA65442.2025.11248010}
}

@Inproceedings{Kim_2020_CVPR_proxy,
author = {Kim, Sungyeon and Kim, Dongwon and Cho, Minsu and Kwak, Suha},
title = {Proxy Anchor Loss for Deep Metric Learning},
booktitle = {Proceedings of the IEEE/CVF Conference on Computer Vision and Pattern Recognition (CVPR)},
month = {June},
year = {2020}
}

@misc{ddnet_wiki,
  title        = {{Official DDraceNetwork Wiki}},
  author       = {{DDNet}},
  howpublished = {\url{https://wiki.ddnet.org/wiki/Main_Page}},
  year         = {2026},
  note         = {Accessed: 2026-01-05},
}

@article{mobilenetsV1,
  title={Mobilenets: Efficient convolutional neural networks for mobile vision applications},
  author={Howard, Andrew G and Zhu, Menglong and Chen, Bo and Kalenichenko, Dmitry and Wang, Weijun and Weyand, Tobias and Andreetto, Marco and Adam, Hartwig},
  journal={ArXiv},
  year={2017}
}

@inproceedings{triplet_margin_loss,
  title={Facenet: A unified embedding for face recognition and clustering},
  author={Schroff, Florian and Kalenichenko, Dmitry and Philbin, James},
  booktitle={Proceedings of the IEEE conference on computer vision and pattern recognition},
  pages={815--823},
  year={2015}
}

@book{game_analytics_book,
title = "Game Analytics: Maximizing the Value of Player Data",
editor = "Anders Drachen and El-Nasr, \{Magy Seif\} and Alessandro Canossa",
year = "2013",
doi = "10.1007/978-1-4471-4769-5",
language = "English",
isbn = "9781447147688",
publisher = "Springer",
edition = "1",
}

@INPROCEEDINGS{factorization_representation,
  author={Sifa, Rafet and Srikanth, Sridev and Drachen, Anders and Ojeda, Cesar and Bauckhage, Christian},
  booktitle={2016 IEEE Conference on Computational Intelligence and Games (CIG)}, 
  title={Predicting Retention in Sandbox Games with Tensor Factorization-based Representation Learning}, 
  year={2016},
  volume={},
  number={},
  pages={1-8},
  doi={10.1109/CIG.2016.7860405}}

@article{deep_metric_learn_survey,
  title={Deep metric learning: A survey},
  author={Kaya, Mahmut and Bilge, Hasan {\c{S}}akir},
  journal={Symmetry},
  volume={11},
  number={9},
  pages={1066},
  year={2019},
  publisher={MDPI}
}

@inproceedings{musgrave2020metric,
  title={A metric learning reality check},
  author={Musgrave, Kevin and Belongie, Serge and Lim, Ser-Nam},
  booktitle={European Conference on Computer Vision},
  pages={681--699},
  year={2020},
  organization={Springer}
}

@INPROCEEDINGS{league_account_sharing,
  author={Da Silva Beserra, Isaac and Camara, Lucas and Da Costa-Abreu, Marjory},
  booktitle={7th International Conference on Imaging for Crime Detection and Prevention (ICDP 2016)}, 
  title={Using keystroke and mouse dynamics for user identification in the online collaborative game league of legends}, 
  year={2016},
  volume={},
  number={},
  pages={1-6},
  doi={10.1049/ic.2016.0076}}

@inproceedings{league_account_hijacking,
author = {Chen, Kuan-Ta and Hong, Li-Wen},
title = {User identification based on game-play activity patterns},
year = {2007},
isbn = {9780980446005},
publisher = {Association for Computing Machinery},
address = {New York, NY, USA},
url = {https://doi.org/10.1145/1326257.1326259},
doi = {10.1145/1326257.1326259},
booktitle = {Proceedings of the 6th ACM SIGCOMM Workshop on Network and System Support for Games},
pages = {7–12},
numpages = {6},
location = {Melbourne, Australia},
series = {NetGames '07}
}

@INPROCEEDINGS{league_smurfing,
  author={Kim, Haerin and Lee, Sangho and Woo, Ji Young and Kim, Huy Kang},
  booktitle={2022 IEEE International Conference on Agents (ICA)}, 
  title={Justice League: Time-series Game Player Pattern Detection to Discover Rank-Skill Mismatch}, 
  year={2022},
  volume={},
  number={},
  pages={42-47},
  doi={10.1109/ICA55837.2022.00014}}

@article{cont_auth_survey,
title = {A study on continuous authentication using a combination of keystroke and mouse biometrics},
journal = {Neurocomputing},
volume = {230},
pages = {1-22},
year = {2017},
issn = {0925-2312},
doi = {https://doi.org/10.1016/j.neucom.2016.11.031},
url = {https://www.sciencedirect.com/science/article/pii/S0925231216314321},
author = {Soumik Mondal and Patrick Bours},
}

@inproceedings{transformers,
 author = {Vaswani, Ashish and Shazeer, Noam and Parmar, Niki and Uszkoreit, Jakob and Jones, Llion and Gomez, Aidan N and Kaiser, \L ukasz and Polosukhin, Illia},
 booktitle = {Advances in Neural Information Processing Systems},
 editor = {I. Guyon and U. Von Luxburg and S. Bengio and H. Wallach and R. Fergus and S. Vishwanathan and R. Garnett},
 pages = {},
 publisher = {Curran Associates, Inc.},
 title = {Attention is All you Need},
 url = {https://proceedings.neurips.cc/paper_files/paper/2017/file/3f5ee243547dee91fbd053c1c4a845aa-Paper.pdf},
 volume = {30},
 year = {2017}
}

@INPROCEEDINGS{csgo_identify,
  author={Zimmer, Franziska and Irvan, Mhd and Perera, Maharage Nisansala Sevwandi and Kobayashi, Ryosuke and Yamaguchi, Rie Shigetomi},
  booktitle={2025 IEEE Conference on Games (CoG)}, 
  title={Fair Play and Identity: In-Game Behavioral Biometrics for Player Identification in Competitive Online Games}, 
  year={2025},
  volume={},
  number={},
  pages={1-8},
  doi={10.1109/CoG64752.2025.11114281}}

@article{beacon,
  title={BEACON: A Multimodal Dataset for Learning Behavioral Fingerprints from Gameplay Data},
  author={Singh, Ishpuneet and Kaur, Gursmeep and Atwal, Uday Pratap Singh and Singh, Guramrit and Singh, Gurjot and Singh, Maninder},
  journal={arXiv preprint arXiv:2605.10867},
  year={2026}
}

@INPROCEEDINGS{trivedi_contrastive_2021,
  author={Trivedi, Chintan and Liapis, Antonios and Yannakakis, Georgios N.},
  booktitle={2021 IEEE Conference on Games (CoG)}, 
  title={Contrastive Learning of Generalized Game Representations}, 
  year={2021},
  volume={},
  number={},
  pages={1-8},
  doi={10.1109/CoG52621.2021.9619107}}

@INPROCEEDINGS{behaviorformer,
  author={Schneider, Tobias and Fahrner, Danial and Antebi, Soren and Uedelhoven, Daniel and Bodden, Lennard and Halscheidt, Sandra and Sifa, Rafet},
  booktitle={2026 IEEE Conference on Games (CoG)}, 
  title={BehaviorFormer: Discovering Behavioral Patterns via Clusterable Next-Token Prediction}, 
  year={2026},
  volume={},
  number={},
  pages={1-4},
  }

\end{document}